# HARMONY SEARCH ALGORITHM FOR THE CONTAINER STORAGE PROBLEM

I. Ayachi [1],[2], R. Kammarti [2]

M. Ksouri [2], P. Borne [1]

[1] LAGIS

Ecole Centrale de Lille, Scientific city - BP 48 - 59651 Villeneuve d'Ascq Cedex - France

ayachiimen@gmail.com , kammarti.ryan@planet.tn

[2] LACS

Ecole Nationale des Ingénieurs de Tunis – BP 37, Le Belvédère, 1002 Tunis – Tunisia

Mekki.Ksouri@insat.rnu.tn, pierre.borne@ec-lille.fr

**ABSTRACT:** *Recently a new metaheuristic called harmony search was developed. It mimics the behaviors of musicians improvising to find the better state harmony. In this paper, this algorithm is described and applied to solve the container storage problem in the harbor. The objective of this problem is to determine a valid containers arrangement, which meets customers' delivery deadlines, reduces the number of container rehandlings and minimizes the ship idle time.*

*In this paper, an adaptation of the harmony search algorithm to the container storage problem is detailed and some experimental results are presented and discussed. The proposed approach was compared to a genetic algorithm previously applied to the same problem and recorded a good results.*

## 1 INTRODUCTION

In recent decades, many efficient algorithms have been developed to solve various optimization problems.
The most of these algorithms are based on linear and nonlinear numerical programming technique which applied the gradient method to the neighborhood of the initial point to improve the solution.

Numerical optimization algorithms are used to solve simple and theoretical models. Nevertheless, there are many complex optimization problems and their resolutions using these algorithms are very difficult. Indeed, if there is more than one local optimum in the problem, the result will depend on the choice of starting point and subsequently the solution is not necessarily the best. In addition, the application of the gradient method becomes difficult and unstable when the objective functions and constraints are numerous.

Due to limits application of the numerical methods, researchers have developed metaheuristics based on simulations to solve complex optimization problems. This metaheuristics combine rules and randomness to imitate natural phenomena. The genetic algorithm is inspired by biological evolutionary process [Goldberg, 1989],[ Holland, 1975]; ant algorithm [Dorigo and al., 1996] and tabu search [Glover, 1977] from animal's behavior; and simulated annealing proposed by [Kirkpatrick and al., 1983] from physical annealing process.

Recently, [Geem and al., 2001] developed a new metaheuristic called Harmony search (HS). The HS algorithm mimics the behaviors of musicians improvising to find a fantastic harmony in terms of aesthetics. [Geem , 2008]

It based on the analogy between music improvisations which seeks the best harmony determined by aesthetic estimation and the searching in optimization process for the optimal solution determined by objective function evaluation [Geem and al., 2001]

The HS algorithm has been applied to various real word optimization problems such as vehicle routing problem [Geem and al., 2005], truss structure design and hydrologic parameter calibration [Lee and Geem, 2004].

Harmony search does not require initial values for the decision variables. It used a stochastic random search based on the probability of the harmony memory considering (HMCR) and pitch adjustment rate (PAR). In addition, this metaheuristic imposes fewer mathematical requirements therefore it can easily be adopted for various types of optimization problems [Lee and Geem, 2004].

In order to demonstrate the performance and the efficiency of HS algorithm, it is applied to a container storage problem which is a classical optimization problem. The container storage problem is classified as a bin packing problem in three dimensions where containers are items and storage spaces in the port are bins used. It falls into the category of NP hard problems.

At each harbour of destination, some containers are unloaded from ship and loaded in the port to be delivered to their customers. Our aim is to determine a valid con-tainers arrangement in the harbor, in order to meet customers' delivery deadlines, reduce the loading/unloading times of these containers as well as their rehandlings number and accordingly to minimize the ship stoppage time.

The main objective of this paper is to determine an optimal solution for containers stowage planning in the port using harmony search technique. Some experimental results are presented to study the influence of containers number and of the harmony memory size on this model.

The rest of this paper is organized as follows: In section 2, a literature review on the container storage problem is presented. Next in section 3, HS algorithm is described. The mathematical formulation of the problem is given in section 4. Then, some experiments and results are presented and discussed, in section 5. Next, a comparative study with the genetic algorithm was performed. Finally, section 7 covers our conclusion.

**2. LITERATURE REVIEW**

In order to be competitive, the port needs to improve its services. In fact, it is important to minimize the ship idle time, which is mostly composed of the loading/unloading containers times. Therefore, it is necessary to make a good arrangement of the containers to import or export in order to be efficiently loaded into the ship, truck, etc.

This work focuses on solving the container storage problem. It consists on affecting, in real-time, containers to available storage spaces in order to minimize the number of containers rehandlings.

Many approaches have been developed to solve this problem: simplified analytical calculations or detailed simulation studies.

Preston and Kozan proposed a genetic algorithm to solve the container location model at seaport terminals. Their objective was to reduce the transfer and the handling time of containers and subsequently the time ships spend at the berth. The results of this approach were compared with the current practice at the port of Brisbane.

In 1999, Kim and Kim proposed a mathematical model for allocating import containers storage area. The strategy of segregation used consists on stacking newly arrived containers on the top of containers that arrived earlier is not allowed. The storage space is allocated in a way of minimizing the number of re-handles.

In order to speed up the loading operation of export containers onto a ship, [Kim and Bea, 1998] developed a mathematical model. To do this, a methodology is proposed to transfer the current yard map for containers into the desirable bay layout. The target of this new layout is to reduce the number of container rehandlings and their travel distance.

Kim an al., in 2000, studied the storage location of an arriving export container in order to minimize the number of relocation movements expected for the loading operation. The most important criteria to consider during the storage location are the configuration of the container stack and the weight container. In this work, a comparative study was provided between resolution via a dynamic programming model and a decision tree.

In [Chen et al, 2004], different metaheuristics (tabu search, simulated annealing and genetic algorithms) were combined to solve the port yard storage optimization problem (PYSOP). The problem is akin to a two dimensional Bin packing problem aims to minimize the space allocated to the cargo within a time interval.

In their work,[Kumar and Vlacic, 2008] presents a simple analytical model for predicting unloading containers times and determining equipment utilization. The prediction model was applied in the Suva's port and has recorded encouraging results.

It is noted that most works studied the storage containers problem, used mathematical and stochastic models. However, these techniques cannot be applied for large scale instances. In addition, they do not take into consideration the dynamic aspect of this problem.

Regarding these limits, it will be more challenging to apply heuristic algorithms that provide good results in a reasonable computation time even for large problem.

In this paper, a recent metaheuristic, the harmony search algorithm is applied to solve the storage containers problem in the port. Our aim is to determine a valid containers arrangement that meet customers' delivery deadlines, reduce the loading/unloading times of these containers as well as their shifting number.

**3. HARMONY SEARCH ALGORITHM**

The musical harmony is improved practice after practice using the set of the pitches played by each instrument. Also, the fitness function is improved iteration by iteration using the values assigned for decision variables. Figure 1 shows this analogy.

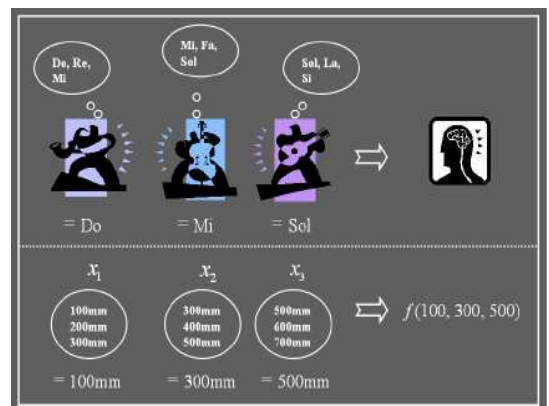

Figure 1: analogy between musical improvisations and optimization process [Geem and al., 2005]

For each music player (saxophonist, double bassist and guitarist) can correspond a decision variable ($x_1$, $x_2$ and $x_3$).

Musical notes list of each instrument (saxophone = {Do, Re, Mi}; double bass = {Mi, Fa, Sol}; and guitar = {Sol, La, Si}) corresponds to the value set of each variable ($x_1$ = {100, 200, 300}; $x_2$ = {300, 400, 500}; and $x_3$ = {500, 600, 700}).

The combination of notes choice (saxophone = {Do}, double bass = {Mi} and guitar = {Sol}) makes a new harmony which will be stored if it is better than other existing harmony.

Similarly, the new solution (100mm, 300mm, 500mm) generated in the optimization process is preserved if it is better than other existing solutions.

The HR algorithm includes five steps: parameters initialization, the harmony memory initialization, new harmony improvisation, memory harmony update and the check of termination criterion.

### 3.1. Parameters initialization

In this step, the optimization problem is specified:

Minimize (or Maximize) $f(\mathbf{x})$; $x_i \in X_i$, $i = 1, 2, \ldots, N$

Where:
- $f(x)$ is an objective function;
- $\mathbf{x}$ is the solution vector composed of decision variables $x_i$;
- $X_i$ is the set of possible values for each decision variable;
- K is the number of possible value for each discrete variable;
- $X_i = \{x_i(1), x_i(2), \ldots, x_i(K)\}$ for discrete variables;
- N is the number of decision variables

The algorithm parameters are also specified during this step such as:
- The harmony memory size (HMS) is the number of solution in the memory.
- The harmony memory considering rate (HMCR): $0 \leq HMCR \leq 1$. His typical values range from 0.7 to 0.99;
- The pitch adjustment rate (PAR): $0 \leq PAR \leq 1$. His selected values range is from 0.1 to 0.5;
- Improvisations number or objective function number.

### 3.2. Harmony memory initialization

During this step, harmony memory showed in equation (1), is randomly generated. Each decision variable ($x_i$) selects a value from its set ($X_i$). Then the fitness values are calculated for the generated solutions.

$$\begin{bmatrix} x_1^1 & x_2^1 & \ldots & x_{N-1}^1 & x_N^1 & f(x^1) \\ x_1^2 & x_2^2 & \ldots & x_{N-1}^2 & x_N^2 & f(x^2) \\ . & \ldots & \ldots & \ldots & \ldots & \ldots \\ . & \ldots & \ldots & \ldots & \ldots & \ldots \\ . & \ldots & \ldots & \ldots & \ldots & \ldots \\ x_1^{HMS-1} & x_2^{HMS-1} & \ldots & x_{N-1}^{HMS-1} & x_N^{HMS-1} & f(x^{HMS-1}) \\ x_1^{HMS} & x_2^{HMS} & \ldots & x_{N-1}^{HMS} & x_N^{HMS} & f(x^{HMS}) \end{bmatrix} \quad (1)$$

### 3.3. New harmony improvisation

Harmony memory is initially crammed; a new harmony vector $x' = (x'_1, x'_2, \ldots, x'_N)$ is generated and compared to existing solutions. This vector is kept if it's better than the worst harmony.

x' is improvised using the following two rules:
- Harmony memory consideration
- Pitch adjustment.

#### 3.3.1. Harmony memory consideration

The value for each decision variable $x'_i$ is randomly chosen using a harmony memory consideration rate (HMCR).

The value of $x'_i$ is selected from any pitches previously stored in HM for this decision variable with a probability of HMCR. While it is chosen with a probability of (1-HMCR) using process described in (2).

$$x'_i \leftarrow \begin{cases} x'_i \in \{x_i^1, x_i^2, \ldots, x_i^{HMS}\} & \text{with probability } HMCR \\ x'_i \in X_i & \text{with probability } (1-HMCR) \end{cases} \quad (2)$$

#### 3.3.2. Pitch adjustment

Each component in the new harmony vector $x' = (x'_1, x'_2, \ldots, x'_N)$ is examined to determine whether it should be adjusted. The variable $x'_i$ will choose a neighboring value with a probability of HMCR×PAR while it kept his original value with a probability of HMCR× (1-PAR).

For example, the note "Do" can be adjusted to "Re" or "Mi" with a probability HMCR×PAR, as shown in Figure 1 and it preserve his pitch value with a probability of HMCR× (1-PAR).

For discrete variable:

$$x'_i(k) \leftarrow \begin{cases} x'_i(k+m) & \text{with probabilty } HMCR \times PAR \\ x'_i(k) & \text{with probabilty } HMCR \times (1-PAR) \end{cases} \quad (3)$$

Where:
k is the index of element in $X_i$
$x'_i(k)$ is the $k^{th}$ element in $X_i$
m is a neighboring index, it's normally +1 or -1.

For continuous variable:

$$x_i' \leftarrow \begin{cases} x_i' \pm bw.\,rand(0,1) & \text{with probability } HMCR \times PAR \\ x_i' & \text{with probability } HMCR \times (1-PAR) \end{cases} \quad (4)$$

where bw is the 'distance bandwidth', the amount of maximum change for pitch adjustment.

### 3.4. Harmony memory update

The new solution is stored in the harmony memory if it's better than the worst of the existing solutions and it respects all problem constraints.

### 3.5. Termination criterion check

Steps (3.3) and (3.4) will repeat while the termination criterion (number of improvisations) is not reached.

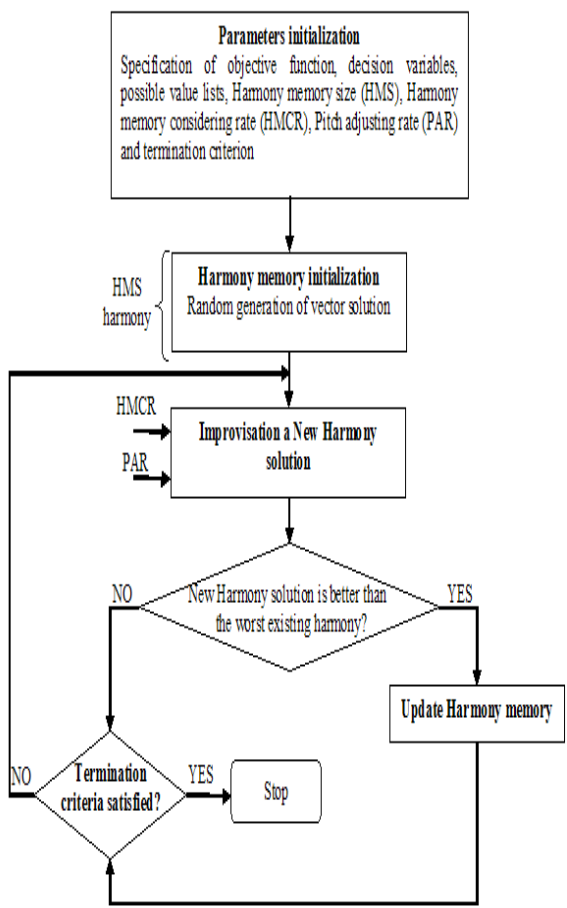

Figure 2: Harmony search algorithm

## 4. PROBLEM FORMULATION

In this section, the proposed approach is detailed by presenting the adopted mathematical formulation and the evolutionary algorithm based on the following assumptions.

### 4.1. Assumptions

In this paper, we suppose that:

- The containers are identical (weight, shape, type) and each is waiting to be delivered to its destination.
- Initially containers are stored at the platform edge or at the vessel.
- A container can be unloaded if all the floor which is above is unloaded
- The containers are loaded from floor to ceiling
- To unload a container, all containers above must be re-handled.

A set of cuboids container localised into a three dimensional cartesian system is showed in the figure 3.

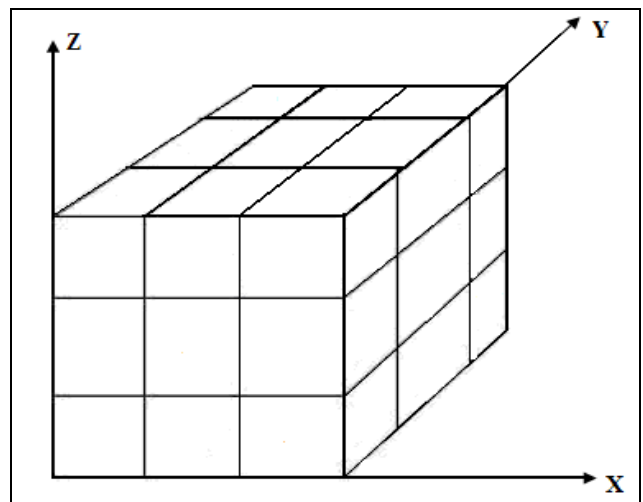

Figure 3: Cartesian coordinate system

### 4.2. Input parameters

Let's consider the following variables:

- i: Container index
- n1: Maximum containers number on the axis X
- n2: Maximum containers number on the axis Y
- n3: Maximum containers number on the axis Z
- $Nc_{floor}$: Maximum containers number per floor, $Nc_{floor}= n1*n2$
- $N_{floor}$: Total number of floors
- $Nc_{floor}(j)$: the containers number in the floor j
- $Nc_{max}$: Maximum containers number, with $Nc_{max}= n1.n2.n3$
- Nc: the containers number
- $N_{iter}$: The iterations number

### 4.3. Mathematical formulation

Let us consider that the space used to stowed containers at the port consisting of a single bay. The fitness function aims to meet customers' delivery date and to reduce the container shifting number. To do that, the following function is used.

Fitness function:

$$\text{Min} \sum_{i=1}^{Nc} P_i m_i X_{i,(x,y,z)}, \forall x = 1..n_1, \forall y = 1..n_2, \forall z = 1..n_3 \quad (5)$$

Where:

$P_i$: Priority value depending on the delivery date $d_i$ of container i to customer, with $P_i = 1/d_i$

$m_i$ : the minimum number of container rehandlings to unload the container i

$x_{i,(x,y,z)}$ is the decision variable,

$$x_{i,(x,y,z)} = \begin{cases} 1, \text{ the container } i \text{ is in the position } (x,y,z) \\ 0 \text{ otherwise} \end{cases} \quad (6)$$

Subject to:

$$Nc_{floor}(j) \geq Nc_{floor}(j+1), \text{ with } j = 1\ldots N_{floor} \quad (7)$$

This constraint equation ensures that a floor lower level contains more containers than directly above.

$$\text{If } x_{i,(x,y,z)} = 0 \text{ then } x_{i,(x,y,z-1)} = 0 \quad (8)$$

The constraint (8) illustrate that a container can only have two positions either on another or on the ground.

### 4.4. Evolution procedure

Here, the evolution procedure used in the proposed approach is detailed.

An initial harmony memory of size HMS is created. The decision variables ($x_i$) represent the possible locations for the containers according to the allocated storage area.

Let's consider the following:

cont = {cont[x][y][z]; $1 \leq x \leq n1$, $1 \leq y \leq n2$, $1 \leq z \leq n3$} which designate the container coordinates.

In this problem, $x_i$ = {cont[0][0][0], cont[1][1][1]... cont[n1-1][n2-1][n3-1]).

The set of possible values ($X_i$) for each decision variable is the container number, $X_i$ = {1, 2, 3... Nc}.
Figure 4 shows an example of harmony representation.

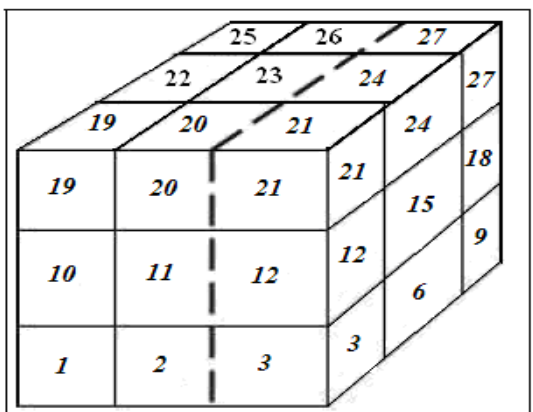

Figure 4: Solution representation

Firstly, the initial harmony memory is randomly generated, where every stored solution must respect all problem constraints (equations (7) and (8)). Figure 5 represents the solution creation algorithm.

```
Begin creat_solution

    For x = 0 to n1-1
        For y = 0 to n2-1
            For z = 0 to n3-1
                Randomly select a container c from the
                ones not already stored
                cont[x][y][z]= c
            End
        End
    End

End
```

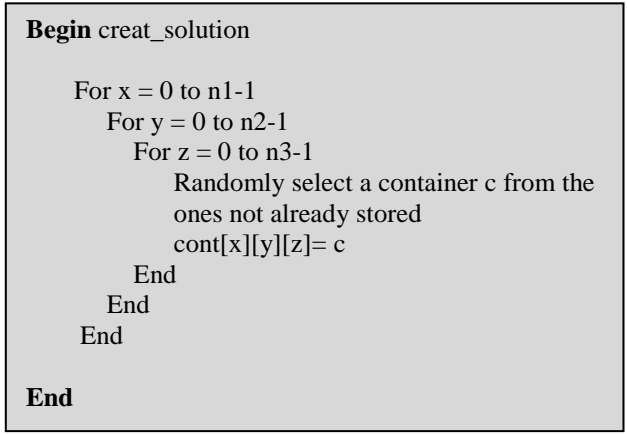

Figure 5: Solution creation algorithm

And then, a new solution is improvised based on the process outlined in section 3.3. This step will be repeated until the termination criterion is satisfied.
The next section describes the experimental results.

## 5. EXPERIMENTAL RESULTS

In this section, different simulations are performed by varying the population size (HMS), the container number (Nc) and the value of the criterion termination. For the proposed approach, the algorithm stops when the solution doesn't improve after $N_{iter}$ iterations.
For the simulation, it is supposed that:
- n1, n2 and n3 will be defined by user, where n1 = n2 = n3.
- The number of containers per harmony is also defined by user.
- The delivery date of each container is randomly generated.
- HMCR= 0.95 and PAR = 0.1

### 5.1 The number of containers influence

To study the influence of the containers number, the algorithm is executed for different values of Nc and each time the best fitness values of the first and the last iterations are given.

Three problem sizes are considered:
- Small sizes (to 64 containers per solution)
- Medium sizes (Between 125 and 750 containers per solution)
- Large sizes (1000 containers per solution).

In this example HMS=50, the stopping criteria (Niter) is set to 20 and each time, the fitness function value is calculated. The results are presented in table 1.

Where $F_i$ is the fitness function value for the best solution in the first iteration and that $F_f$ is the fitness function value for the best solution in the last iteration.

As it can be seen, the value of $F_f$ improves considerably regarding the one of $F_i$.

| Nc | Fi | $F_f$ |
|---|---|---|
| 64 | 73.58 | 50.98 |
| 125 | 163.61 | 139.08 |
| 343 | 769.25 | 574.43 |
| 729 | 1831.26 | 1610.66 |
| 1000 | 2511.79 | 2390.60 |

Table 1: Evolution of the fitness function according to the containers number

### 5.2 The stopping criteria value influence

In order to study the influence of the stopping criteria value, we varied $N_{iter}$ and fixed the container number (Nc=64) and the size of the harmony memory (HMS=50).

| $N_{iter}$ | Fi | $F_f$ |
|---|---|---|
| 20 | 73.58 | 50.98 |
| 50 | 78.13 | 41.46 |
| 100 | 75.26 | 38.39 |
| 150 | 77.03 | 37.75 |
| 175 | 81.40 | 37.63 |
| 200 | 70.49 | 37.50 |

Table 2: The influence of generation number

According to the results illustrated in Table 2, we note that higher is the value of the stopping criteria, better is the quality of the fitness function. However, the execution time increases with the stopping criteria value.

### 5.3 The harmony memory size influence

Through this example, the size of the problem is fixed to 125 containers and $N_{iter}$ to 100 iterations and the number of solutions (HMS) is varied to study his influence on the algorithm behaviour. The results are presented in the table 3.

| HMS | $F_i$ | $F_f$ |
|---|---|---|
| 20 | 182.83 | 106.42 |
| 50 | 195.90 | 102.86 |
| 75 | 189.53 | 101.26 |
| 100 | 168.63 | 99.82 |
| 125 | 207.55 | 99.82 |

Table 3: Evolution of the fitness values according to the harmony memory size

The results shown in the table 3 indicate that higher is the harmony memory size, better is the value of the fitness function.

### 6. COMPARATIVE STUDY:

To evaluate the results generated by the proposed HS algorithm, a comparative study with the genetic algorithm, proposed by Kammarti and al., was performed. This GA can be described as follows: Initially, a first generation is randomly generated. Then, a two-point crossover operator is performed to two parent selected using the roulette-wheel method. The mutation operator consists of permuting two randomly selected containers. To compare these two approaches, we vary the containers numbers. The results are given by Figure 6.

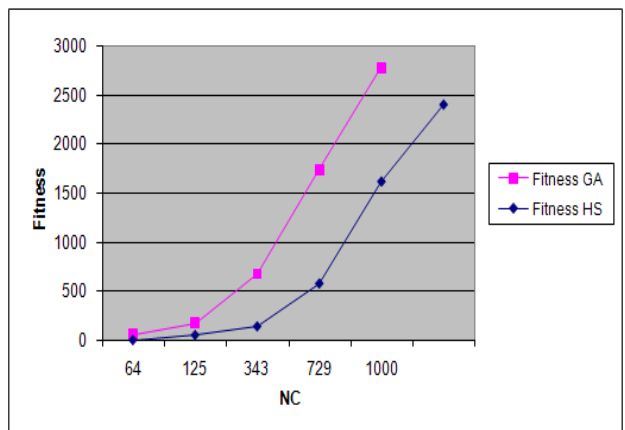

Figure 6: Evolution of the fitness values generated by the HS and AG, according the containers number

As it can be seen, the fitness value generated by the HS algorithm is better for all problem sizes.

In addition, we set the number of containers to 125 and the iteration number to 100 and we varied the population size. The results presented in Table 4 indicate that the HS algorithm generate better results.

| Population size | Fitness (GA) | Fitness (HS) |
|---|---|---|
| 20 | 144.14 | 106.42 |
| 50 | 124.65 | 102.86 |
| 75 | 121.53 | 101.26 |
| 100 | 115.87 | 99.82 |
| 125 | 107.36 | 99.82 |

Table 4: Effects of population size on computational performance of GA and HS

### 7. SUMMARY AND CONCLUSIONS

In this paper, a new metaheuristic inspired by music improvisations, harmony search, is applied to solve the containers storage problem at the port. The objective aims to determine the best containers arrangement that meet customers' delivery dates and reduce the number of container rehandlings.

These results were compared to other works treating the same problem using other metaheuristics, as [Kammarti and al., 2009] and recorded good results.

The proposed approach has provided encouraging results and seems to have a potential to be successfully applied to more difficult variants of the containers storage problem, e.g. the one when the containers are of different types and sizes.

**REFERENCES**

Chen, P., Z. Fu, A. Lim, and B. Rodrigues, 2004, Port yard storage optimization, *IEEE Transactions on Automation Science and Engineering*.

Dorigo, M., V. Maniezzo and A. Colorni, 1996, The ant system: optimization by a colony of cooperating agents, *IEEE Trans. Systems Man Cybernet Part B*, Vol. 26, No. 1, pp. 2942.

Geem, Z.W., J.H. Kim and G.V. Loganathan, 2001, A new heuristic optimization algorithm: harmony search, *Simulation 76*, p. 60-68.

Geem, Z.W., 2008, Harmony Search Applications in Industry, *Soft Computing Applications in Industry 2008.* p. 117-134

Geem, Z.W., K.S. Lee and Y. Park, 2005, Application of Harmony Search to Vehicle Routing, *Journal of Applied Sciences*, p. 1552-1557

Goldberg, D.E., 1989, Genetic Algorithms in search Optimization and Machine Learning. *Addison Wesley, MA*.

Holland, J.H., 1975. *Adaptation in Natural and artificial Systems*, University of Michigan Press, Ann Arbor, MI.

Kammarti, R., I. Ayachi, M. Ksouri and P. Borne, 2009, Evolutionary Approach for the Containers Bin-Packing Problem, *Studies in Informatics and Control.*

Kim, K. H. and H.B. Kim, 1999, Segregating space allocation models for container inventories in port container terminals, *International Journal of Production Economics*, 59: 415–423.

Kim, K. H. and J. W. Bae, 1998, Re-marshaling export containers in port container terminals, *Computers & Industrial Engineering*, 35(3-4): 655–658.

Kim, K. H., Y. M. Park, and K. R. Ryu, 2000, Deriving decision rules to locate export containers in container yards. *European Journal of Operational Research*.

Kirkpatrick, S., C.D. Gelatt and M.P. Vecchi, 1983, Optimization by simulated annealing, *Science* 220 (4598), p. 671–680.

Kumar, S. and L. Vlacic, 2008, Performance Analysis of Container Unloading Operations at the Port of Suva Using a Simplified Analytical Model (SAM), *Journal of Advanced Computational Intelligence and Intelligent Informatics*, Vol.12 No.4

Lee, K.S., and Z.W Geem, 2004, A new structural optimization method based on the harmony search algorithm. *Computers and Structures*, 82: 781-798.

Preston, P. and E. Kozan, 2001, An approach to determine storage locations of containers at seaport terminals, *Computers & Operations Research*, 983-995